\documentclass[conference]{IEEEtran}
\IEEEoverridecommandlockouts
\usepackage{cite}
\usepackage{url}
\usepackage{amsmath,amssymb,amsfonts}
\usepackage{algorithmic}
\usepackage{graphicx}
\usepackage{textcomp}
\usepackage{xcolor}
\def\BibTeX{{\rm B\kern-.05em{\sc i\kern-.025em b}\kern-.08em
    T\kern-.1667em\lower.7ex\hbox{E}\kern-.125emX}}
\begin{document}

\title{Enhancing Large Vision-Language Models with Layout Modality for Table Question Answering on Japanese Annual Securities Reports}


\author{\IEEEauthorblockN{1\textsuperscript{st} Hayato Aida}
\IEEEauthorblockA{\textit{Stockmark} \\
Tokyo, Japan \\
hayato.aida@stockmark.co.jp}
\and
\IEEEauthorblockN{2\textsuperscript{nd} Kosuke Takahashi}
\IEEEauthorblockA{\textit{Stockmark} \\
Tokyo, Japan \\
kosuke.takahashi@stockmark.co.jp}
\and
\IEEEauthorblockN{3\textsuperscript{rd} Takahiro Omi}
\IEEEauthorblockA{\textit{Stockmark} \\
Tokyo, Japan \\
takahiro.omi@stockmark.co.jp}
}

\maketitle

\begin{abstract}
With recent advancements in Large Language Models (LLMs) and growing interest in retrieval-augmented generation (RAG), the ability to understand table structures has become increasingly important.  
This is especially critical in financial domains such as securities reports, where highly accurate question answering (QA) over tables is required.  
However, tables exist in various formats—including HTML, images, and plain text—making it difficult to preserve and extract structural information.  
Therefore, multimodal LLMs are essential for robust and general-purpose table understanding.  
Despite their promise, current Large Vision-Language Models (LVLMs), which are major representatives of multimodal LLMs, still face challenges in accurately understanding characters and their spatial relationships within documents.

In this study, we propose a method to enhance LVLM-based table understanding by incorporating in-table textual content and layout features.  
Experimental results demonstrate that these auxiliary modalities significantly improve performance, enabling robust interpretation of complex document layouts without relying on explicitly structured input formats.
The TableCellQA dataset—including rendered images, layout (bounding‐box) annotations, and QA data—will be publicly released upon publication (license confirmation in progress).
\end{abstract}
\begin{IEEEkeywords}
Table QA, VQA, LVLM, Multimodal, Layout
\end{IEEEkeywords}

\section{Introduction}

Tables are structured representations of data and are widely used in business documents such as reports, spreadsheets, and financial statements.  
In particular, financial documents like annual securities reports often contain complex tables that are essential for quantitative decision-making.  
The NTCIR-18 U4 task \cite{ntcir18-u4-overview} provides a valuable benchmark, offering tables extracted from Japanese securities reports in structured HTML format.

However, real-world tables appear in a variety of formats—including HTML, Markdown, CSV, and scanned PDFs—many of which lack explicit structural annotations.  
This diversity poses significant challenges for automated table understanding, especially when accurate value extraction is required for downstream tasks such as financial question answering (QA).  
Recent advances in Large Vision-Language Models (LVLMs) have enabled joint processing of image and text inputs, offering a promising path toward format-agnostic table interpretation.

To better evaluate table understanding in such multimodal settings, we reformulate the original NTCIR-18 U4 Table QA dataset into a simplified version that focuses purely on direct value extraction from table cells.  
We refer to this task as \textbf{TableCellQA}.  
Unlike the original task—which includes deriving answers through arithmetic operations or transformations—TableCellQA requires models to identify and extract exact cell values based solely on structural and semantic alignment with the question.  
This setup allows us to isolate and assess the model's ability to comprehend table structure without external reasoning.

In this study, we propose a multimodal framework that enhances LVLMs by incorporating not only the image modality, but also text and layout information.  
We decompose each HTML table into three distinct modalities—image, text, and layout—and feed them into a modified LVLM to improve its structural understanding capabilities.

Experimental results demonstrate that incorporating layout and text modalities leads to more than a 7\% improvement in accuracy over the image-only baseline.  
Ablation studies further highlight the critical role of these auxiliary modalities in correctly identifying relevant cells, providing new insights into effective multimodal strategies for table understanding in real-world business documents.

\section{Related Work}
Visual document understanding encompasses the extraction and interpretation of information from document images to answer relevant queries. 
Within the area of visual document understanding, Table QA specifically focuses on comprehending tabular information contained within documents. 
Various benchmarks have been established for assessing performance of downstream tasks in this field. 
For instance, DocVQA \cite{docvqa} involves extracting and understanding textual and visual content from diverse document images to respond accurately to posed questions. 
Similarly, datasets like CORD \cite{park2019cord} and FUNSD \cite{jaume2019funsddatasetformunderstanding} focus on specialized tasks such as receipt understanding and form information extraction from scanned documents, respectively. 
These datasets leverage multimodal information, including visual features, textual content, and spatial layouts, which are crucial elements for accurate table comprehension.

Recent advancements in transformer-based architectures have further improved multimodal document understanding. LayoutLMv3 \cite{layoutlmv3}, for example, incorporates visual, textual, and spatial modalities to achieve state-of-the-art results across various document understanding tasks.

More recently, LVLMs have demonstrated their potential for document understanding tasks by leveraging the strong text-processing capabilities of high-performing LLMs. 
Models such as LLaVA \cite{llava} and LLaVA-OneVision \cite{llava-ov} have shown promising capabilities in general multimodal tasks. 
Specifically, Qwen2-VL \cite{qwen2-vl} has achieved state-of-the-art performance on the DocVQA benchmark.

Meanwhile, Zheng et al.~\cite{zheng2024multimodaltableunderstanding} proposed Table-LLaVA, a multimodal vision-language model specifically optimized for table understanding from images. 
They reformulated table understanding as an instruction-following problem and constructed the MMTab dataset to support this paradigm. 
While Table-LLaVA primarily operates on table images without direct access to underlying structured representations like HTML or CSV, it demonstrates that combining visual information with task-specific instructions can substantially enhance table reasoning capabilities.

While models like Table-LLaVA focus solely on visual inputs, some prior approaches have explored incorporating additional modalities such as text and layout information.

LayoutLM\cite{layoutlm}, on the other hand, is specifically tailored for document image understanding. 
It extends traditional encoder-based language models by incorporating not only textual content but also the spatial layout of documents. 
Based on the Transformer architecture, LayoutLM integrates token embeddings with 2D positional embeddings that represent the coordinates of text within a document. This approach allows the model to capture the structural information essential for understanding complex documents, such as forms or receipts. 
Subsequent versions, like LayoutLMv2\cite{layoutlmv2} and LayoutLMv3\cite{layoutlmv3}, further enhance this capability by incorporating actual image embeddings alongside text and layout information, enabling a more comprehensive understanding of documents that include both textual and visual elements.

Additionally, models that leverage only textual and layout information have also been explored. 
LayTextLLM\cite{laytextllm} focuses on integrating textual content with spatial layout information. 
It achieves this integration by mapping each bounding box to a single embedding and interleaving it with the corresponding text. 
This approach efficiently addresses sequence length issues and leverages the autoregressive traits of LLMs, enabling effective document understanding without relying on explicit visual inputs. 

With the advancement of LLMs utilizing Transformer decoders with layout awareness, QA datasets incorporating layout information have also been proposed. LayoutLLM\cite{layoutllm} introduces a QA dataset that leverages document images along with the text and layout information within them, while also presenting a baseline architecture.
The construction of this dataset has the potential to enhance LLMs’ ability to comprehensively understand text, layout, and images.

Motivated by these developments, we build upon existing LVLM frameworks by: (1) adopting an LVLM capable of efficiently processing high-resolution images and (2) extending its architecture to incorporate comprehensive multimodal inputs—including image, text, and layout information—specifically optimized for the task of table understanding.

\section{Methods}

\subsection{Task Definition: TableCellQA}

The goal of TableCellQA is to evaluate a model’s ability to extract precise cell values from tables in response to natural language questions.  
This task is derived from the Table QA subtask in NTCIR-18 U4, which originally required arithmetic reasoning and unit interpretation (e.g., converting from thousands to millions of yen).  
While valuable, such requirements introduce additional complexity that can obscure the model’s true table understanding capabilities.

To isolate and measure this core skill, we reformulate the original task into TableCellQA.  
In TableCellQA, the answer to each question is defined as the raw value of a single cell that aligns with the question’s intent.  
We utilize the cell IDs provided in the original dataset, which specify the location of the answer cell, to replace answers that originally involved computation with the corresponding raw cell values.  
This setting allows us to focus purely on structural and semantic comprehension without external reasoning steps.

\begin{figure}
    \centering
    \includegraphics[width=1\linewidth]{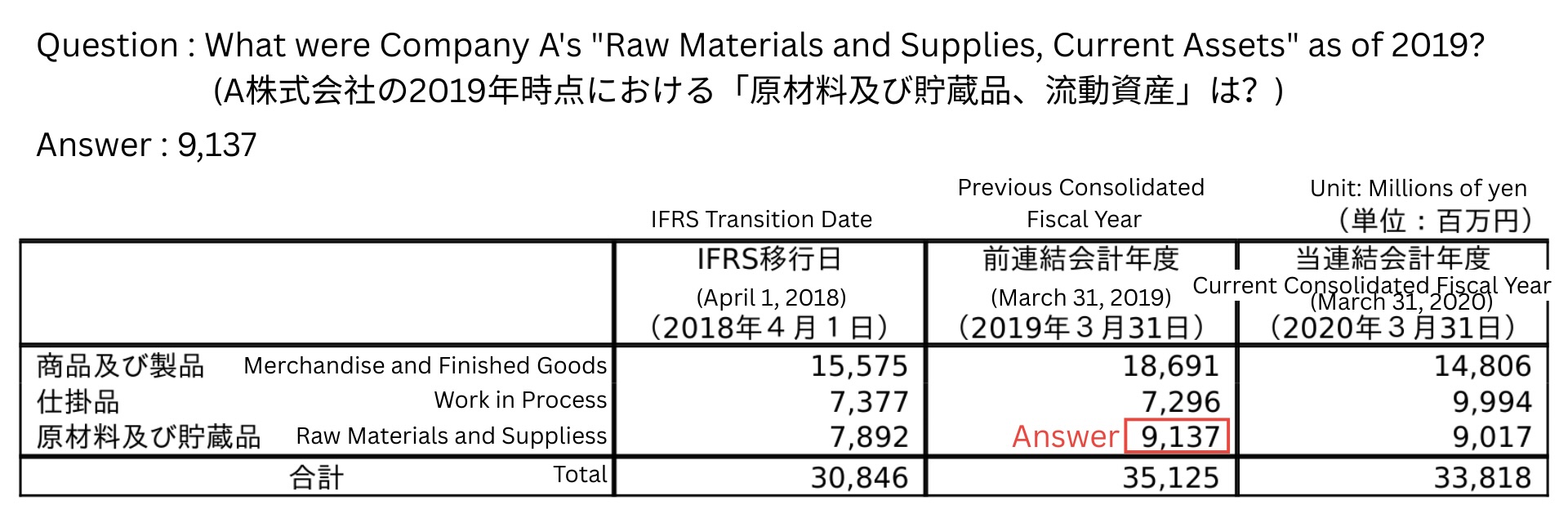}
    \caption{Example of TableCellQA}
    \label{fig:qa_sample}
\end{figure}

Figure~\ref{fig:qa_sample} shows an example from TableCellQA, where the model must identify the correct cell in the table and extract its value without performing any additional reasoning.

\subsection{Data Preparation}

\begin{figure}
    \centering
    \includegraphics[width=1\linewidth]{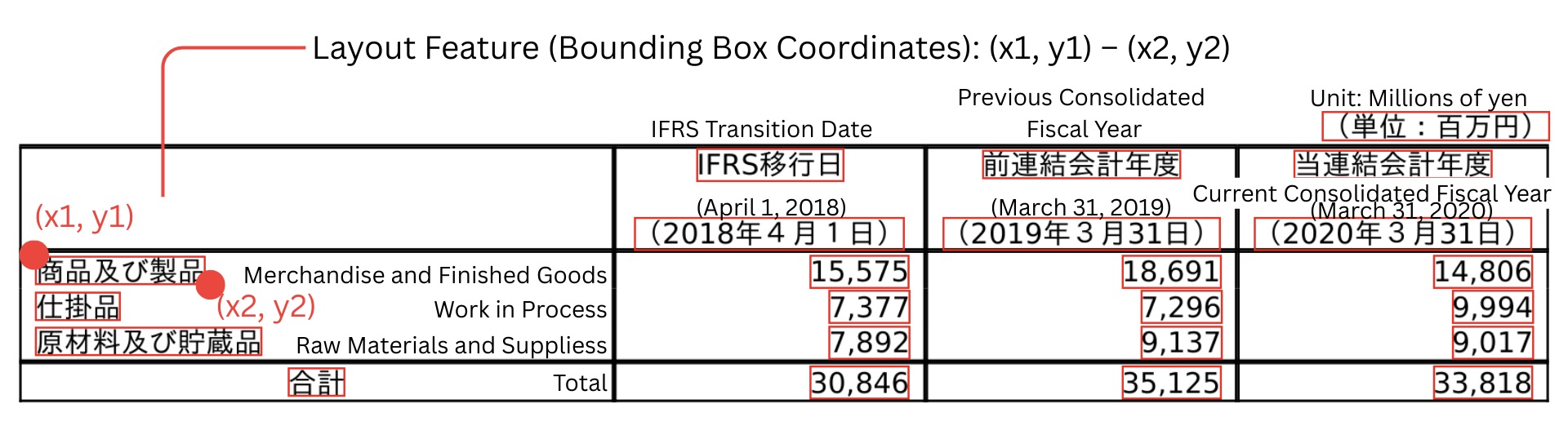}
    \caption{Layout feature example. Each text region is represented by a bounding box $(x1, y1)$–$(x2, y2)$ indicating its spatial position in the table image.}
    \label{fig:example_layout}
\end{figure}

\begin{figure}
    \centering
    \includegraphics[width=1\linewidth]{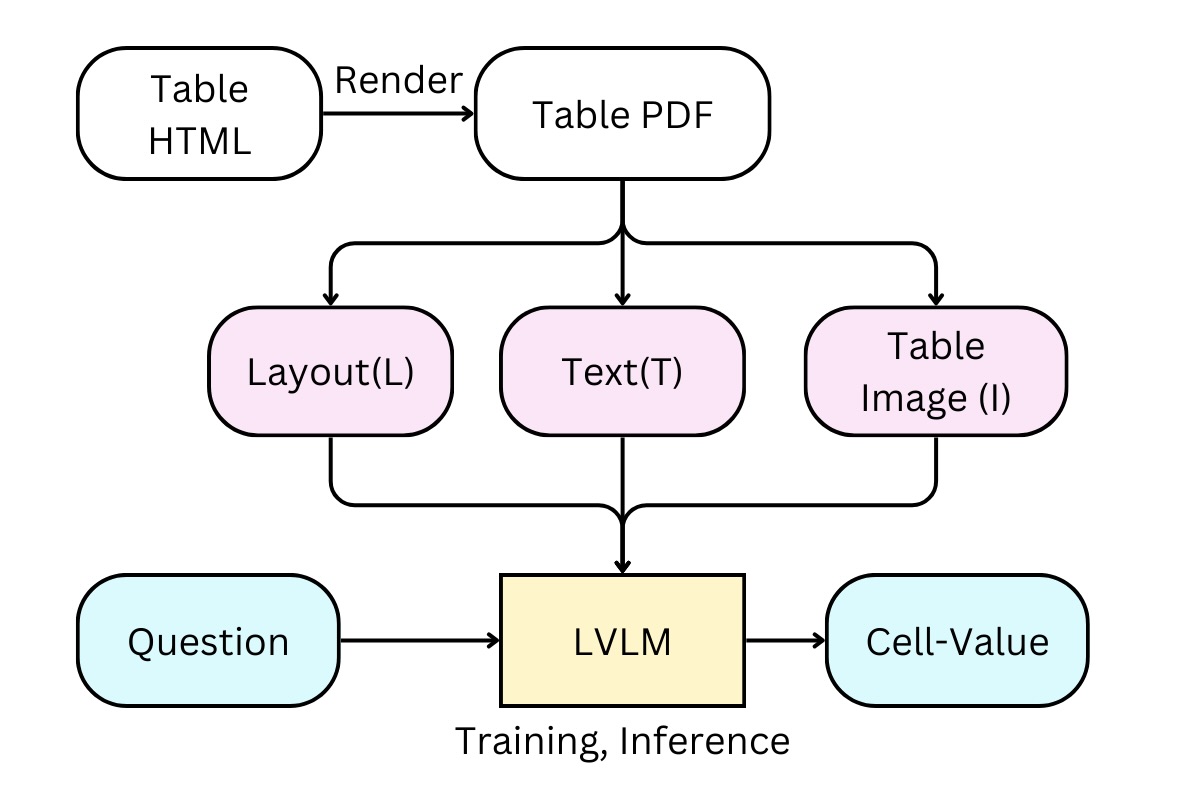}
    \caption{Overall system architecture}
    \label{fig:system_architecture}
\end{figure}

Figure~\ref{fig:system_architecture} illustrates the processing pipeline used to obtain layout (L), text (T), and image (I) modalities from HTML tables.  
To generate the image and layout modalities, we first rendered HTML tables as PDFs.  
From these PDFs, we extracted the layout modality as bounding-box coordinates representing the position of each text span, along with the corresponding OCR-extracted text and table images.  
A concrete example of the layout extraction is shown in Figure~\ref{fig:example_layout}.

For constructing the QA data for TableCellQA, we mapped each question in the NTCIR-18 U4 Table QA dataset to its corresponding answer cell using the provided cell IDs, and retrieved the raw cell value as the new answer.  
Specifically, we parsed the HTML tables to extract a dictionary that maps cell IDs to their corresponding values from the HTML attributes, which was used for converting the original QA pairs into TableCellQA format.

This conversion was applied consistently to both the training and test splits.  
By replacing computation-involved answers with direct cell values, we created instruction-style QA pairs that emphasize table structure comprehension.

In addition, we prepared structured table text data by extracting clean HTML representations (removing non-structural tags and attributes) and converting them into Markdown and JSON formats to explore the effect of different text styles.

After filtering out cases where table rendering failed, we obtained a final dataset consisting of 10,278 training examples and 1,303 test examples.  
This dataset serves as the foundation for evaluating our proposed TableCellQA task under various modality configurations.


\subsection{Model Construction}
LVLMs typically accept only images as input. 
However, we modify the architecture to incorporate text and layout modalities in addition to images for this task.

Following previous studies such as LayTextLLM \cite{laytextllm}, layout embedding is achieved by converting bounding box coordinates into the hidden dimensions of an LLM using a two-layer MLP. 

\begin{figure}
    \centering
    \includegraphics[width=1\linewidth]{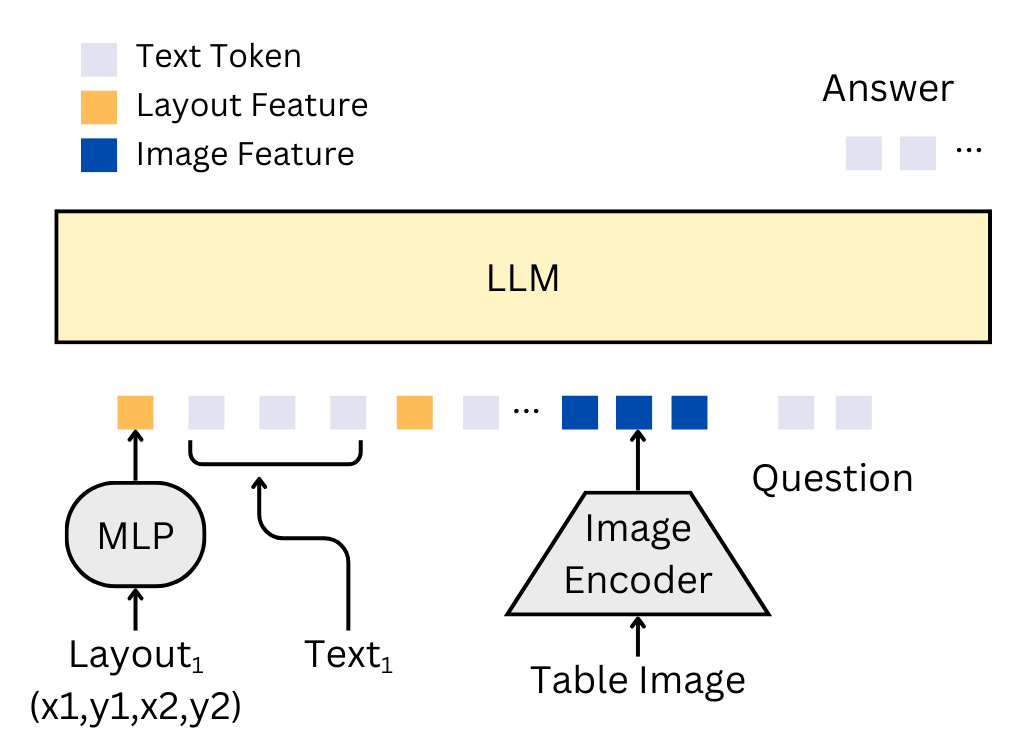}
    \caption{The architecture of our LVLM. Each text token is paired with its corresponding layout feature (e.g., Layout$_1$ and Text$_1$), where the layout is represented by bounding box coordinates. Layout features are encoded via an MLP and combined with text and image features as input to the LLM.}
    \label{fig:model_architecture}
\end{figure}

Each bounding box is represented as a 4-dimensional vector $\mathbf{b} = (x_{\text{min}}, y_{\text{min}}, x_{\text{max}}, y_{\text{max}})$, where $(x_{\text{min}}, y_{\text{min}})$ and $(x_{\text{max}}, y_{\text{max}})$ denote the top-left and bottom-right coordinates of the bounding box, respectively.  
The layout embedding $\mathbf{e}_{\text{L}} \in \mathbb{R}^d$ is then obtained via a two-layer multilayer perceptron (MLP):
\[
\mathbf{e}_{\text{L}} = \text{MLP}(\mathbf{b})
\]
where $\text{MLP}$ denotes a feed-forward neural network with non-linear activations, and $d$ is the embedding dimension.
Each layout embedding is treated as a single token input to the LLM and paired with the corresponding text, as shown in Figure \ref{fig:model_architecture}. This approach enables the model to process text while maintaining its spatial correspondence within the table.

Each text span within the table is associated with a layout embedding \(\mathbf{e}_{L,i}\) and a text embedding \(\mathbf{e}_{T,i}\).  
These embeddings are concatenated to form a combined representation for each span:
\[
\mathbf{h}_i = [\mathbf{e}_{L,i}; \mathbf{e}_{T,i}]
\]
The combined sequence of all spans is denoted as:
\[
H_{L+T} = (\mathbf{h}_1, \mathbf{h}_2, \ldots, \mathbf{h}_N)
\]
where \(N\) is the number of text spans in the table.  
Finally, the model input \(\mathbf{X}\) is constructed by concatenating the image embedding, the question embedding, and the sequence \(H_{L+T}\):
\[
\mathbf{X} = [H_{L+T}; \mathbf{e}_{\text{image}}; \mathbf{e}_{\text{question}}]
\]
This combined input \(\mathbf{X}\) is then fed into the LLM for processing.

In the experiments, different combinations of image, layout, and text inputs are tested to analyze the contributions of each modality. 
To ensure consistency in analysis, all evaluations are conducted under the same conditions, allowing for a direct comparison of the impact of each input type.

\subsection{Evaluation Metrics}

We evaluated model performance using two complementary metrics: Accuracy and Approximate Normalized Levenshtein Similarity (ANLS)~\cite{anls2019scenetextvisualquestion}.  
Accuracy measures the proportion of predictions that exactly match the ground-truth cell values, providing a strict evaluation of model precision.  
In addition, we report ANLS, which quantifies the character-level similarity between predicted and reference values.  
This allows for a more tolerant evaluation that accounts for minor recognition errors, such as small typos or OCR noise, which are common when relying solely on the image modality in conventional vision-language models (VLMs).


\section{Experiments}
In this section, we present the experimental conditions and results using the data and models described in the previous section.

\subsection{Training Conditions}
Table~\ref{tab:experiments_list} summarizes the input modalities evaluated.  
L+T+I denotes models using layout, text, and image jointly; L+T uses layout and text only; T+I uses text and image; L+I uses layout and image; and I uses image only.  
In addition, models were trained and evaluated using structured table representations in HTML, Markdown, and JSON formats for comparison with traditional text-based table understanding approaches.

All fine-tuning experiments were conducted based on the llava-onevision-qwen2-7b-ov (LLaVA-ov-7b) model~\cite{llava-ov}.  
Training for TableCellQA was performed for 2 epochs with a batch size of 8, a learning rate of 1e-5, and a warmup ratio of 0.03.

\begin{table}
    \centering
    \caption{Description of Each Input Modality}
    \begin{tabular}{ll}
    \hline
    Modality or Format  & Description \\
    \hline
    L+T+I & Layout, Text, and Image \\
    L+T   & Layout and Text \\
    T+I   & Text and Image \\
    L+I   & Layout and Image \\
    I     & Image only \\
    \hline
    HTML     & Cleaned HTML text extracted from the document \\
    Markdown & Rendered Markdown converted from HTML \\
    JSON     & JSON representation generated from HTML \\
    \hline
    \end{tabular}
    \label{tab:experiments_list}
\end{table}

\subsection{Results}
Table~\ref{tab:comparison_modalities} presents the evaluation results across different input modality combinations.  
We also include comparisons with the zero-shot performance of strong pretrained multimodal models.  
The "FT" column indicates whether the model was fine-tuned on the TableCellQA dataset.

\begin{table}[t]
  \centering
  \caption{Comparison of Modalities (TableCellQA-Test)}
  \label{tab:comparison_modalities}
  \begin{tabular}{lclrr}
  \hline
  Model   & FT & Modality & Acc. & ANLS \\
  \hline
  LLaVA-ov-7b  & Yes &  L+T+I & 0.9478 & \textbf{0.9666} \\
  LLaVA-ov-7b  & Yes &  L+T   & \textbf{0.9509} & 0.9645 \\
  LLaVA-ov-7b  & Yes &  T+I   & 0.9394 & 0.9571 \\
  LLaVA-ov-7b  & Yes &  L+I   & 0.8733 & 0.9129 \\
  LLaVA-ov-7b  & Yes &  I     & 0.8764 & 0.9139 \\
  \hline
  Qwen2.5-VL-72B & No & I & 0.6631 & 0.7610 \\
  GPT-4o         & No & I & 0.5748 & 0.6708 \\
  \hline
  \end{tabular}
\end{table}

\begin{table}[t]
  \centering
  \caption{Comparison of Structured Table Text (TableCellQA-Test)}
  \label{tab:comparison_structured_table}
  \begin{tabular}{lclrr}
  \hline
  Model   & FT & Format & Acc. & ANLS \\
  \hline
  LLaVA-ov-7b  & Yes & Markdown   & 0.9540 & 0.9657 \\
  LLaVA-ov-7b  & Yes & JSON       & 0.9563 & 0.9693 \\
  LLaVA-ov-7b  & Yes & HTML       & \textbf{0.9655} & \textbf{0.9762} \\
  \hline
  \end{tabular}
\end{table}

To compare the zero-shot performance of high-capacity multimodal models, we additionally conducted inference using Qwen2.5-VL-72B~\cite{qwen2-vl, qwen2.5-VL} and GPT-4o (gpt-4o-2024-08-06)~\cite{openai2024gpt4o}.

\subsubsection{Comparison of Text, Layout and Image Modalities}
To investigate the effectiveness of incorporating layout and in-image text information in LVLMs, we compared various combinations of the I (image), T (text), and L (layout) modalities.  
The combination of text (T) and layout (L) achieved the highest Accuracy, while the combination of all three modalities (L+T+I) yielded the highest ANLS score.  
Performance dropped in all cases where either text or layout information was removed. In particular, the L+I setting, which excludes text, showed the most significant performance degradation. The T+I setting, which lacks layout information, also underperformed in both Accuracy and ANLS compared to the T+L setting, where only the image modality was excluded.  

These results indicate the critical importance of both textual content and its spatial layout for multimodal table understanding. Among the modalities, text contributes the most to performance, followed by layout information.  
Interestingly, while the combination of layout and text (L+T) achieved the highest Accuracy, the addition of the image modality (L+T+I) led to a higher ANLS score.  
This trend suggests that incorporating image information, despite introducing minor OCR errors, provides complementary evidence that improves character-level similarity metrics.  
Conversely, the highest Accuracy under L+T highlights the advantage of relying solely on clean text embeddings aligned with layout information, without introducing redundant visual representations where the textual content is already explicitly provided.  
However, it should be noted that these differences are within a margin of less than 0.5\%, and may fall within the natural variance of the dataset.


To clarify the effect of layout information, we compared the inference results of the L+T+I and I+T settings and analyzed cases where the model failed when layout information was excluded.  

\begin{figure}
    \centering
    \includegraphics[width=1\linewidth]{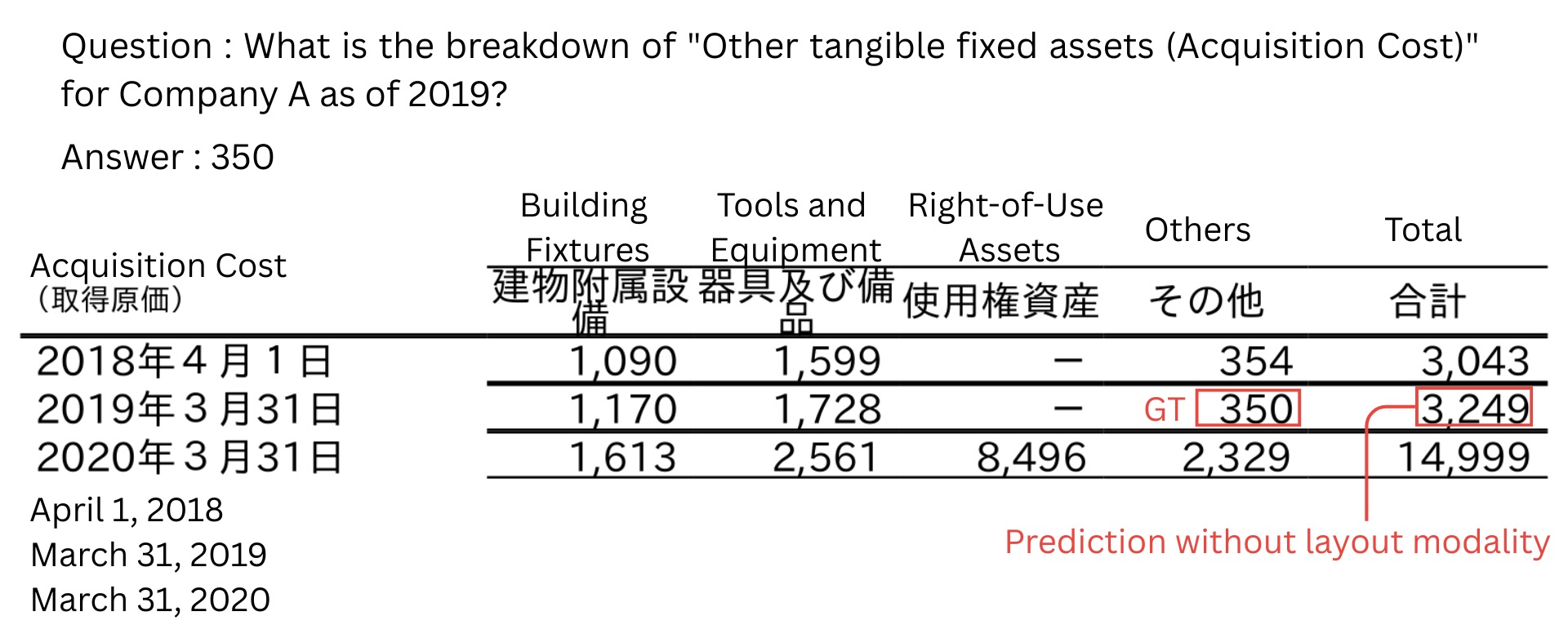}
    \caption{An example of a prediction error caused by the absence of layout modality.(T+I)}
    \label{fig:layout_qa_sample}
\end{figure}

This phenomenon is illustrated in Figure~\ref{fig:layout_qa_sample}, where the model incorrectly selects neighboring cells due to the absence of spatial layout information. Since models using only text input are not provided with positional relationships beyond sequence order, such errors are likely due to a lack of spatial context. These findings suggest that layout information plays a crucial role in recognizing the row and column structures of tables.

\begin{figure}
    \centering
    \includegraphics[width=1\linewidth]{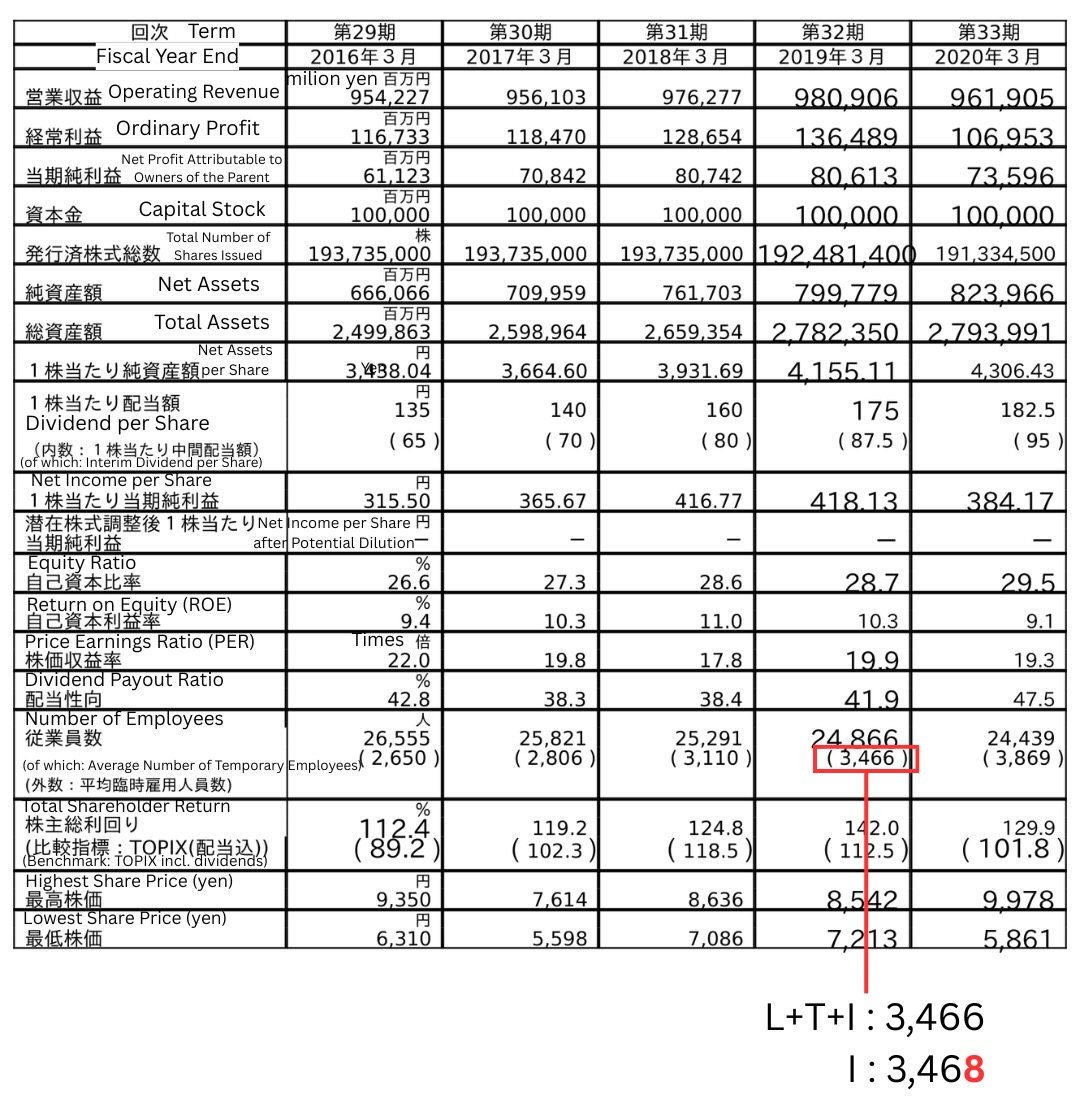}
    \caption{An example of a large table where OCR-related errors occur. While the L+T+I setting correctly predicts the answer "3,466", the image-only (I) setting produces an incorrect answer "3,468", highlighting the limitations of visual-only approaches.}
    \label{fig:large_table}
\end{figure}


Performance deteriorated significantly under the image-only condition, particularly for larger tables with smaller and denser textual elements.
This performance degradation stems from the limitations of current LVLM OCR capabilities, which struggle to accurately recognize text in small-scale or visually crowded contexts.
For instance, in large tables with dense text, we observed basic OCR errors, such as predicting "3,468" instead of the correct value "3,466," illustrating the susceptibility of image-only models to minor recognition mistakes (see Figure~\ref{fig:large_table}).
These findings highlight the need for not only enhanced OCR capabilities but also the incorporation of additional modalities, such as layout and text features, to reliably parse complex visual data.

Overall, the results underscore that textual content remains the primary source of information for accurate table understanding, with layout information providing crucial spatial context.  
While image information currently plays a complementary role and may introduce minor noise, it also offers valuable support, particularly for tables with rich visual elements, suggesting promising potential for future extensions in multimodal table understanding.

\subsubsection{Comparison of Structured Table Text}
Table~\ref{tab:comparison_structured_table} presents the evaluation results for different formats of structured table text.
Structured table text represented in HTML, JSON, and Markdown formats significantly outperformed all settings that utilized image or layout modalities, highlighting the effectiveness of clean and explicitly structured data for table understanding tasks. This suggests that properly structured tables allow LLMs to perform at their best. Among the three formats, HTML achieved the highest performance.  

Markdown tables do not support cell merging, and merged cells are instead duplicated to create a pseudo-structure during conversion. This leads to a loss of structural information compared to the original HTML, which likely contributed to the lowest performance among the three formats.  

JSON-formatted tables, on the other hand, can preserve merged cell information and are structurally equivalent to HTML in terms of information content. However, representing tables in JSON is less common in real-world applications, and it is expected that the underlying models are pre-trained on a larger volume of HTML tables than JSON ones. These factors likely account for the performance differences observed among the structured table text formats.

Although structured HTML tables demonstrated the highest performance in our experiments, such structured table text is typically not available in real-world scenarios, where business documents are commonly provided in PDF format. In practice, an additional module would be required to infer or reconstruct structured table text from unstructured sources, and the final performance would depend heavily on the accuracy of that module.  

Moreover, extending this approach to documents containing figures or photographs remains a challenge. In such cases, our proposed multimodal approach incorporating layout, text, and image modalities remains applicable. Thus, while structured table text is highly effective when available, multimodal approaches remain essential for handling unstructured, visually complex documents encountered in practical applications.

In our dataset, the HTML tables used as the rendering source retain complete structural information and serve as an upper bound for table understanding performance.
Notably, the T+L+I setting—combining image, text, and layout—achieves results that are remarkably close to those of the structured formats.
This indicates that the multimodal representation effectively captures the underlying table structure, and can serve as a practical substitute when explicit structural annotations are unavailable.


\subsubsection{Performance Comparison with State-of-the-Art Models}

We compared the performance on a visual question answering task using only table images, employing Qwen2.5-VL-72B and GPT-4o.
The zero-shot performance of these state-of-the-art models was lower than that of all the methods proposed in this study.
In particular, there was a notable performance gap compared to models fine-tuned even with image-only inputs, indicating that task-specific adaptation and familiarity with dataset characteristics play a critical role in achieving high accuracy.
Interestingly, the zero-shot performance of GPT-4o was lower than that of Qwen2.5-VL.
Qwen2.5-VL is reported to have been trained on a large number of synthetic document images generated from HTML sources, which may contribute to its strong performance on table-like visual inputs.
In contrast, GPT-4o is likely trained with a greater emphasis on general-purpose capabilities.
These results highlight the importance of pre-training data distribution in tasks that require domain-specific visual understanding, such as table-based reasoning.


\subsubsection{Comparison of Training Data}
Since the proposed layout modality is not present in existing LVLMs, pre-training is conducted to help the model adapt to this format. 
The LayoutLLM-SFT dataset \cite{layoutllm} is designed for document-based QA tasks and includes OCR text and bounding-box coordinates alongside images and QA data. 
To explore the impact of data augmentation for learning layout features, we conducted an additional experiment in which 50\% of the LayoutLLM-SFT dataset was used for pre-training. This experiment was conducted only on models that utilize the L+T+I modalities.

Table \ref{tab:comparison_data} shows the comparison results with and without pre-training on the LayoutLLM-SFT dataset.  
When pre-training was conducted using LayoutLLM-SFT, both Accuracy and ANLS scores decreased.  
This finding suggests that the inherent capability of a general-purpose LVLM to comprehend structured visual information may already be sufficient—or even preferable—for TableCellQA tasks, compared to specialized pre-training that could unintentionally limit the model's generalization ability.

\begin{table}[t]
  \centering
  \caption{Comparison of Training Data}
  \label{tab:comparison_data}
  \begin{tabular}{lrr}
    \hline
     Model   & Acc. & ANLS \\
    \hline
  L+T+I               & 0.9478 & 0.9666  \\
  L+T+I w LayoutLLM-SFT & 0.9355 & 0.9544 \\
  \hline
  \end{tabular}
\end{table}

\section{Conclusions}

In this study, we developed models for table question answering that leverage multimodal information, including text, images, and layout.  
We found that incorporating textual and layout information within tables significantly enhances the table understanding capability of LVLMs.  
Among the three modalities—text, layout, and image—we observed that text contributes the most to performance, followed by layout information, and then image information.

On the other hand, structured table text achieved the highest overall performance, reaffirming the importance of explicitly defined table structure.  
Our approach can be seen as an intermediate solution between end-to-end table understanding from raw images and methods that rely on fully structured HTML-based tables.  
This approach allows us to bypass the need for full table structure reconstruction while mitigating the limitations of image-only understanding, thereby enabling more generalizable table comprehension.

We also found that task-specific pretraining did not necessarily improve performance, suggesting that preserving the generalization ability of LVLMs is crucial for flexible and robust table understanding across diverse document formats.

In future work, we aim to extend our method beyond text-only tables to handle more complex documents that contain mixtures of text, images, and figures.  
Ultimately, we seek to develop multimodal LLMs that are well-suited for real-world business scenarios.


\bibliographystyle{plain}
\bibliography{conference_041818}

\end{document}